\documentclass[10pt,twocolumn,letterpaper]{article}
\usepackage{algorithm}
\usepackage{algpseudocode}
\usepackage{cvpr}
\usepackage{times}
\usepackage{epsfig}
\usepackage{graphicx}
\usepackage{subfigure}
\usepackage{amsmath}
\usepackage{amssymb}
\usepackage{multirow}
\usepackage{booktabs}
\usepackage{mathrsfs}
\usepackage{tabu}
\usepackage{float}
\usepackage{subfigure}
\usepackage{multirow}
\usepackage{array}
\usepackage{enumerate}
\usepackage{booktabs}
\usepackage[T1]{fontenc}
\usepackage[utf8]{inputenc}
\usepackage{authblk}
\usepackage{color}
\usepackage{fancyhdr}

\usepackage[breaklinks=true,bookmarks=false]{hyperref}

\cvprfinalcopy %

\begin{document}
\pagestyle{empty}

\title{Circulant Binary Convolutional Networks: Enhancing the Performance of 1-bit DCNNs with  Circulant Back Propagation }

\author{\textcolor[rgb]{1,1,1}{jironrongg}Chunlei~Liu,\textsuperscript{1} Wenrui~Ding,\textsuperscript{2} Xin~Xia,\textsuperscript{1}  Baochang~Zhang,\textsuperscript{4}\thanks{Baochang Zhang is the corresponding author.}  \textcolor[rgb]{1,1,1}{j} Jiaxin~Gu,\textsuperscript{4}\textcolor[rgb]{1,1,1}{jiergrong}
	{  Jianzhuang~Liu,\textsuperscript{3} Rongrong~Ji,\textsuperscript{5,6} David Doermann\textsuperscript{7}}  \\
\textsuperscript{1} School of Electronic and Information Engineering, Beihang University,\\
\textsuperscript{2} Unmanned System Research Institute, Beihang University,
\textsuperscript{3} Huawei Noah's Ark Lab,\\
\textsuperscript{4} School of Automation Science and Electrical Engineering, Beihang University,\\
\textsuperscript{5} School of Information Science and Engineering, Xiamen University,\\
\textsuperscript{6} Peng Cheng Laboratory,
\textsuperscript{7} University at Buffalo \\
\{liuchunlei, ding, xiaxin, bczhang\}@buaa.edu.cn\\
}

\maketitle

\maketitle
\thispagestyle{empty}

\begin{abstract}
 The rapidly decreasing computation and memory cost  has recently driven the success of many applications in the field of deep learning. Practical applications of deep learning in resource-limited hardware, such as embedded devices and smart phones, however, remain challenging. For binary convolutional networks, the reason lies in  the degraded  representation  caused by  binarizing full-precision filters.
	To address this problem, we propose new  circulant  filters (CiFs) and a circulant binary convolution (CBConv)  to enhance the capacity of binarized convolutional  features  via our  circulant back propagation (CBP). The CiFs can be easily incorporated into existing  deep convolutional neural networks (DCNNs), which leads to  new  Circulant Binary Convolutional Networks (CBCNs). Extensive experiments confirm that the performance gap between the 1-bit and full-precision DCNNs is minimized by increasing the filter diversity, which further increases the representational ability in our networks.
Our experiments on ImageNet show that CBCNs achieve 61.4\% top-1 accuracy with ResNet18.
Compared to the state-of-the-art such as XNOR, CBCNs can achieve up to 10\% higher top-1 accuracy with more powerful representational ability.
\end{abstract}

\section{Introduction}
Deep convolutional neural networks (DCNNs) have been successfully demonstrated on many computer vision tasks such as object detection and image classification.  DCNNs deployed in practical environments, however, still face many challenges.  It is particularly true when the portability  and real time performance are required. This is critical  because models of vision applications can require very large memory, making them impractical for most embedded platforms. Besides, floating-point inputs and network weights along with forward or backward data flow can result in a significant computational burden.
\begin{figure}[!t]
\centering
\includegraphics[scale=0.8]{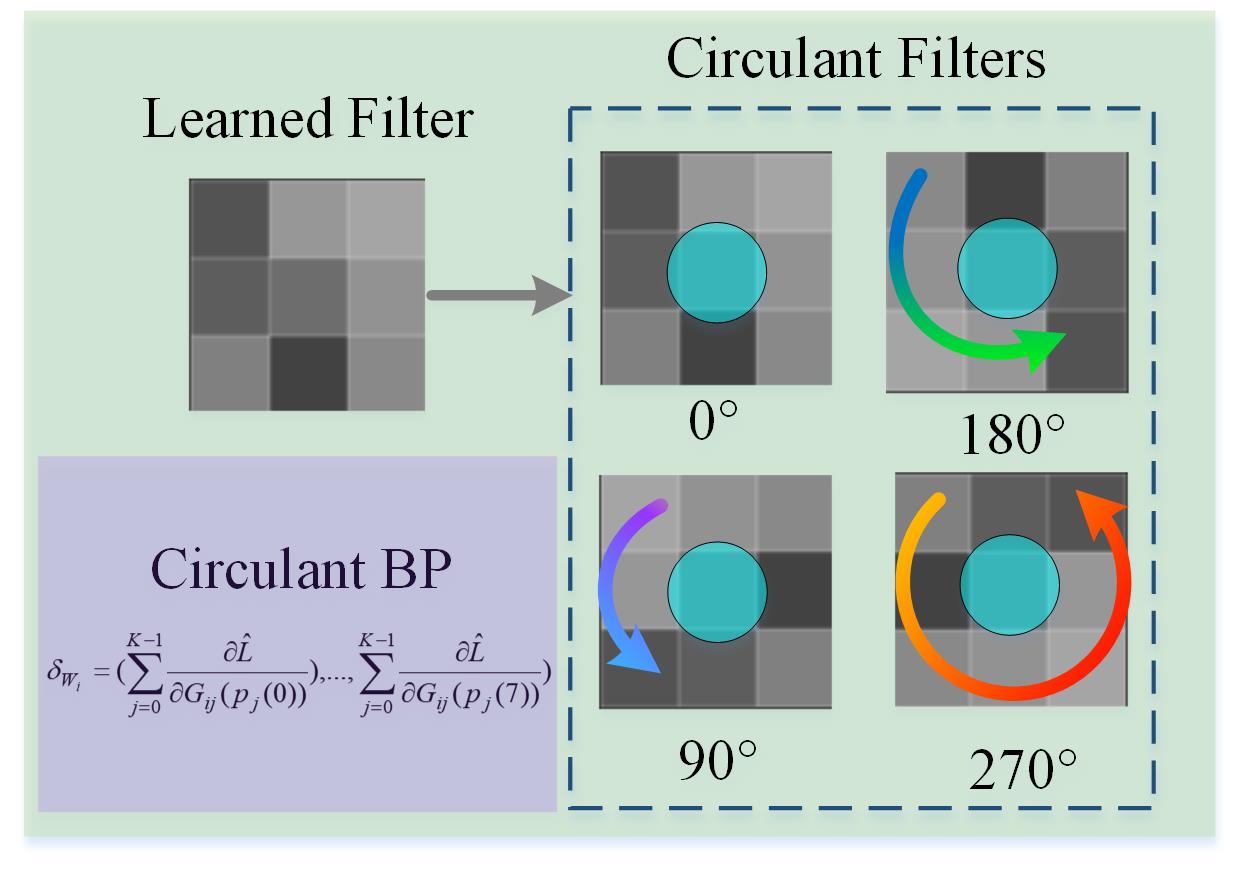}
\caption{Circulant back propagation (CBP).
We manipulate the learned convolution filters using the circulant transfer matrix, which is employed to build our CBP. By doing so, the capacity of the binarized convolutional features are significantly enhanced, e.g., robustness to the orientation variations in objects,  and the performance gap between the 1-bit and full-precision DCNNs is minimized. In the example, 4 CiFs are produced based on the learned filter and the circular matrix.
}
\label{main}
\end{figure}
\begin{figure*}[!t]
\centering
\includegraphics[scale=0.4]{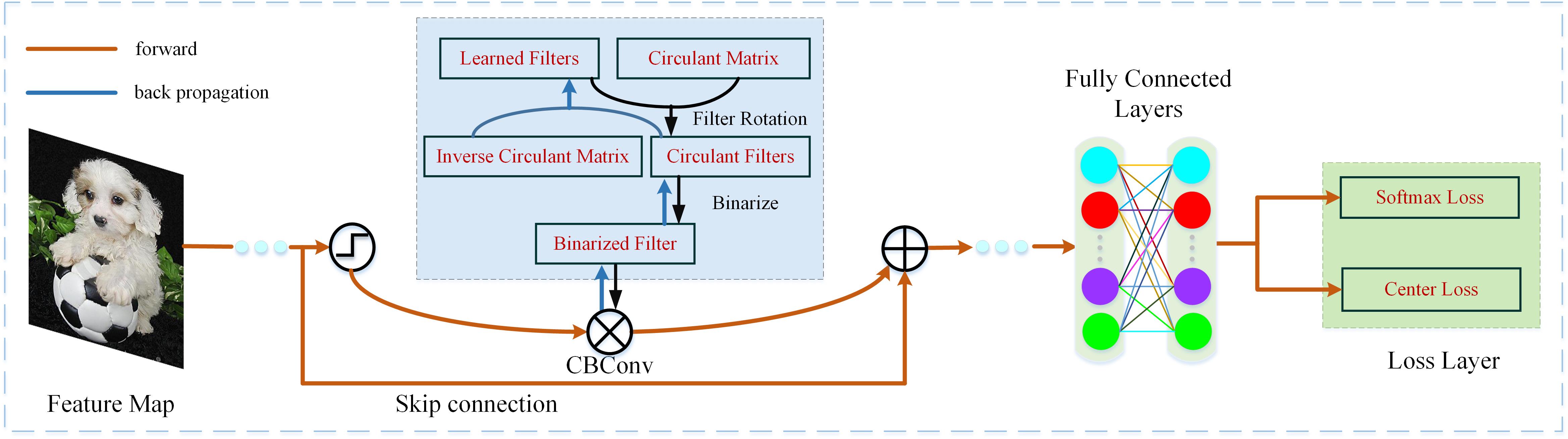}
\caption{ Circulant Binary Convolutional Networks (CBCNs) are designed based on circulant and binary filters to variate the orientations of the learned filters in order to increase the representational ability.
By considering the center loss and softmax loss in a unified framework, we achieve much better performance than state-of-the-art binarized models.
Most importantly, our CBCNs also achieve the performance comparable to well-known full-precision ResNets and WideResNets.
The  circulant binary filters are only shown for demonstrating the computation procedure, which are not saved for testing. }
\label{forwardbp}
\end{figure*}

Binary filters instead of using full-precision weights have been investigated in DCNNs to compress the deep models to handle the aforementioned problems.
They are also called 1-bit DCNNs, as each weight parameter and activation can be represented by a single bit.
As presented in \cite{Rastegari2016XNOR}, XNOR has both the convolution weights and inputs attached to the convolution be approximated with binary values, providing an efficient implementation of convolutional operations, particularly by reconstructing the unbinarized filters with a single scaling factor.
More recently, Bi-Real Net \cite{Liu2018Bi} explores a new variant of residual structure to preserve the real activations before the sign function and TBN \cite{Wan2018TBN} replaces the sign function with a threshold-based ternary function to obtain ternary input tensor. Both provide an optimal tradeoff among memory, efficiency and performance. A warm-restart learning-rate schedule in \cite{Mark2018Training} is adopted to accelerate the training for 1-bit-per-weight networks. Furthermore, a method called WAGE \cite{Wu2018Training} is proposed to discretize both training and inference, where not only weights and activations but also gradients and errors are quantized. In these previous methods, however, the binarization of the filters often degrades the representational ability of the models  for the  rotation variations in objects. %

Inspired by Oriented Response Networks \cite{Zhou2017ORN} which already show their powerful ability in
within-class rotation-invariance, we also employ similar way to enhance the representative ability which destoryed by the binarization process. To the best of our knowledge, we are the first to use the idea of orientation in binary network with a asynchronous way, opening up a promising direction for pursuing efficient networks based on virtual shared filter banks.
In this paper, we introduce  circulant filters (CiFs) and the circulant binary convolution (CBConv) to actively calculate  diverse  feature maps, which can improve the representational ability of the resulting binarized DCNNs. The key insight of producing  CiFs to help back propagation is shown in Fig.\;\ref{main}.
Compared to previous binarized DCNN filters, CiFs are  defined based on a  circulant operation on each learned filter. A new  circulant back propagation (CBP) algorithm is also introduced to develop an end-to-end trainable DCNN.
During the convolution, CiFs are used to produce diverse feature maps which provide the binarized DCNNs with the ability to capture variations previously unseen.
Instead of introducing extra functional modules or new network topologies, our method implements CBConv onto the most basic element of DCNNs, the convolution operator.
Thus, it can be naturally fused with modern DCNN architectures, upgrading them to more expressive and compact  Circulant Binary Convolutional Networks (CBCNs) for resource limited applications.
We design a simple and unique variation process, which is deployed at each layer and can be solved within the same pipeline of the new CBP algorithm.
In addition, we  consider the center loss to further enhance the performance of CBCNs as shown in Fig.\;\ref{forwardbp}.
Thanks to the low model complexity, such an architecture is less prone to over-fitting and suitable for resource-constrained environments.
Our CBCNs reduce the required storage space of full-precision models by a factor of 32, while achieving better performance than existing binarized filters based DCNNs.
The contributions of this paper are summarized as follows:

(1) CiFs are used to obtain more robust feature representation, e.g.,  orientation variations in objects,  which minimize the performance gap between full-precision DCNNs and binarized DCNNs.

(2) We develop a CBP algorithm to reduce the loss during back propagation and make convolutional networks more compact and efficient.
Experimental results show that CBP is not only effective, but also converged quickly.

(3) The presented circulant convolution is generic, and can be easily used on existing DCNNs, such as ResNets and conventional DCNNs.
Our highly compressed models outperform state-of-the-art binarized models by a large margin  on MNIST, CIFAR and ImageNet databases.

\begin{table*}[]
\centering
\caption{A brief description of variables and operators used in the paper.}
\begin{tabular}{llllllll}
\hline
$M:$circulant transfer matrix  & {$G:$ circulant filter}&$W:$ learned filter &$F:$ feature map &     \\
${\delta}_W:$ the gradient of the learned filter $W$&$P:$  inverse circulant transfer matrix& $\hat{G}:$ binarized filter&$L$: loss function&\\\hline
$K:$ number of orientations  for each filter       & {$i:$ filter index} & $j:$ orientation index   &$l:$ layer index&   \\
$g:$ input feature map index & $h:$ output feature map index&${\eta}:$ learning rate&&\\\hline
\end{tabular}
\label{notation}
\end{table*}

\section{Related Work}

DCNNs with a large number of parameters consume considerable computational resources. From our practical study, about half of the power consumption of a DCNN is related to the model size.
Many attempts have been made to accelerate and simplify DCNNs while avoiding the increase of the errors. Network binarization is one of the most popular approaches, which is briefly reviewed below.

The research in \cite{Lai2017Deep} demonstrates that the storage of real-valued deep neural networks such as AlexNet \cite{Krizhevsky2012ImageNet}, GoogLeNet \cite{Szegedy2014Going} and VGG-16 \cite{Simonyan2014Very} can be reduced significantly when their 32-bit parameters are quantized to 1-bit.
Expectation BackPropagation (EBP) in \cite{Soudry2014Expectation} uses a variational Bayesian approach to achive high performance with a network with binary weights and activations. BinaryConnect (BC) \cite{Courbariaux2015BinaryConnect} extends the idea of EBP by employing 1-bit precision weights (1 and -1).
Later, Courbariaux et al. \cite{Courbariaux2016Binarized} propose BinaryNet (BN) that is an extension of BC and further constrains activations to +1 and -1, binaring the input (except the first layer) and the output of each layer.
BC and BN both achieve sufficiently high accuracy on small datasets such as MNIST, CIFAR10 and SVHN.
According to Rastegari et al. \cite{Rigamonti2013Learning}, BC and BN are not very successful on large-scale data sets.
XNOR \cite{Rastegari2016XNOR} has a different binarization method and a network architecture.
Both the weights and inputs attached to the convolution are approximated with binary values, which results in an efficient implementation of the  convolutional operations by reconstructing the unbinarized filters with a single scaling factor.
Recent studies such as MCN \cite{Wang2018Modulated} and Bi-Real Net \cite{Liu2018Bi} have been conducted to explore new network structures and training techniques for binarizing both weights and activations while minimizing accuracy degradation using a concept similar to XNOR. MCN introduces modulated filters to recover the unbinarized filters and leads to a new architecture to calculate the network model. Bi-Real Net connects the real activations to the activations of consecutive blocks through an identity shortcut.

The results of these studies are encouraging, but due to the weight binarization process, the representational ability of the networks can be degraded. This inspires us to seek a way to increase the filter variations in order to increase the network representation ability. In particular, for the first time, we use the circulant matrix to build CiFs for our binarized CNNs. We also develop a CBP algorithm to make the DCNNs more compact and effective in an end-to-end framework.

\section{Methodology}
We design a specific architecture in CBCNs based on CiFs, and train it with a new BP algorithm.
Attempting to  increase the representational ability reduced by the binarization process, CiFs are designed to enrich the binarized filters for the enhancement of the network performance.
As shown in the experiments, the performance drop is marginal even when the learned network parameters are highly compressed. First of all, Table \ref{notation} gives the main notation used in this paper.
\begin{figure}[!t]
\centering
\includegraphics[scale=1.2]{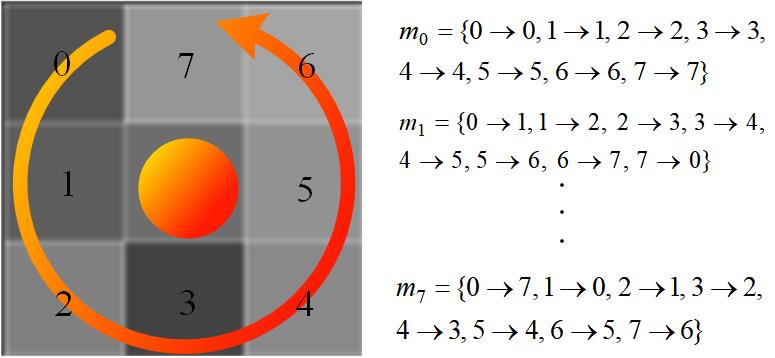}
\caption{Illustration of the circulant transfer matrix $M$ for $K=8$.
The center position stays unchanged, and the remaining numbers are circled in a counter-clockwise direction.
Each column of $M$ is obtained from $m_0$ with a rotation angle $\in\{0^{\circ}, 45^{\circ},..., 315^{\circ}\}$.
It clearly shows that a circulant filter explicitly encodes the position and orientation.
}
\label{cifmatrix}
\end{figure}

\subsection{ Circulant Transfer Matrix $M$}

A circulant matrix $M$ is defined by a single vector in the first column, with cyclic permutations of the vector with offset equal to the column index in the remaining columns. An important property of the circulant matrix is that it can produce different representations using simple vectors or matrices.
With this unique characteristic, we define the  circulant transfer matrix of $K$ columns as $M=(m_0,...,m_j,...,m_{K-1})$, $j=\{0,1,..,K-1\}$:
\begin{equation}
\begin{aligned}
M=\left(\begin{matrix}
&\text{0}~~~~\text{7}~~~~\text{6}~~~~\text{5}~~~~\text{4}~~~~\text{3}~~~~\text{2}~~~~\text{1}~~~ \\
& \text{1}~~~~\text{0}~~~~\text{7}~~~~\text{6}~~~~\text{5}~~~~\text{4}~~~~\text{3}~~~~\text{2}~~~\\
&\text{2}~~~~\text{1}~~~~\text{0}~~~~\text{7}~~~~\text{6}~~~~\text{5}~~~~\text{4}~~~~\text{3}~~~\\
&\text{3}~~~~\text{2}~~~~\text{1}~~~~\text{0}~~~~\text{7}~~~~\text{6}~~~~\text{5}~~~~\text{4}~~~\\
&\text{4}~~~~\text{3}~~~~\text{2}~~~~\text{1}~~~~\text{0}~~~~\text{7}~~~~\text{6}~~~~\text{5}~~~\\
&\text{5}~~~~\text{4}~~~~\text{3}~~~~\text{2}~~~~\text{1}~~~~\text{0}~~~~\text{7}~~~~\text{6}~~~\\
&\text{6}~~~~\text{5}~~~~\text{4}~~~~\text{3}~~~~\text{2}~~~~\text{1}~~~~\text{0}~~~~\text{7}~~~\\
&\text{7}~~~~\text{6}~~~~\text{5}~~~~\text{4}~~~~\text{3}~~~~\text{2}~~~~\text{1}~~~~\text{0}~~~\\
\end{matrix} \right),
\end{aligned}
\label{M}
\end{equation}
where $K=8$ and $8$ vector rotations are used to form ${M}$. The first column $m_0$ corresponds to the numbers in Fig. \ref{cifmatrix}, and the other columns are obtained by a counter-clockwise rotation of the numbers. Each column of $M$ represents one rotation angle $\in$ $\{0^{\circ}$, $45^{\circ}$, $90^{\circ}$, $135^{\circ}$, $180^{\circ}$, $225^{\circ}$, $270^{\circ}$, $315^{\circ}$\}. We set $m_0$ to correspond to the learned filter and $m_{1-7}$ to the derived rotated versions of $m_0$.

\subsection{Circulant Filters (CiFs)}

\begin{figure}
  \centering
  \subfigure[Traditional convolution]{
    \label{4_1} %
    \includegraphics[width=3.3in]{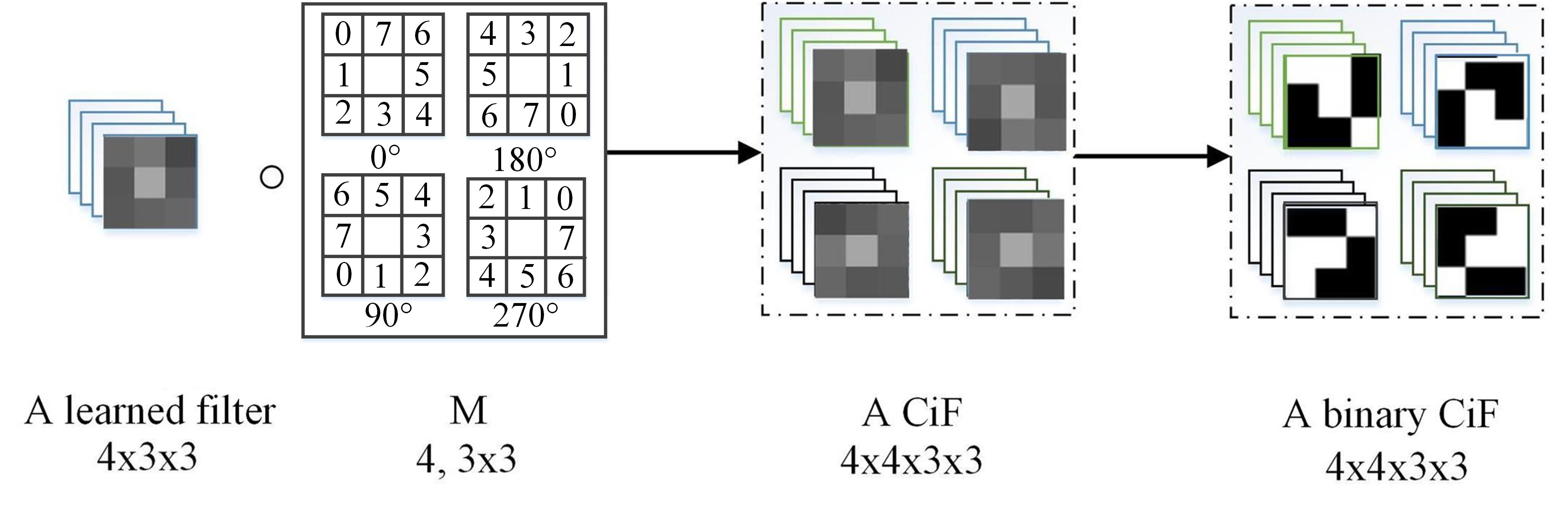}
  }
  \subfigure[CBConv]{
    \label{4_2} %
    \includegraphics[width=3.3in]{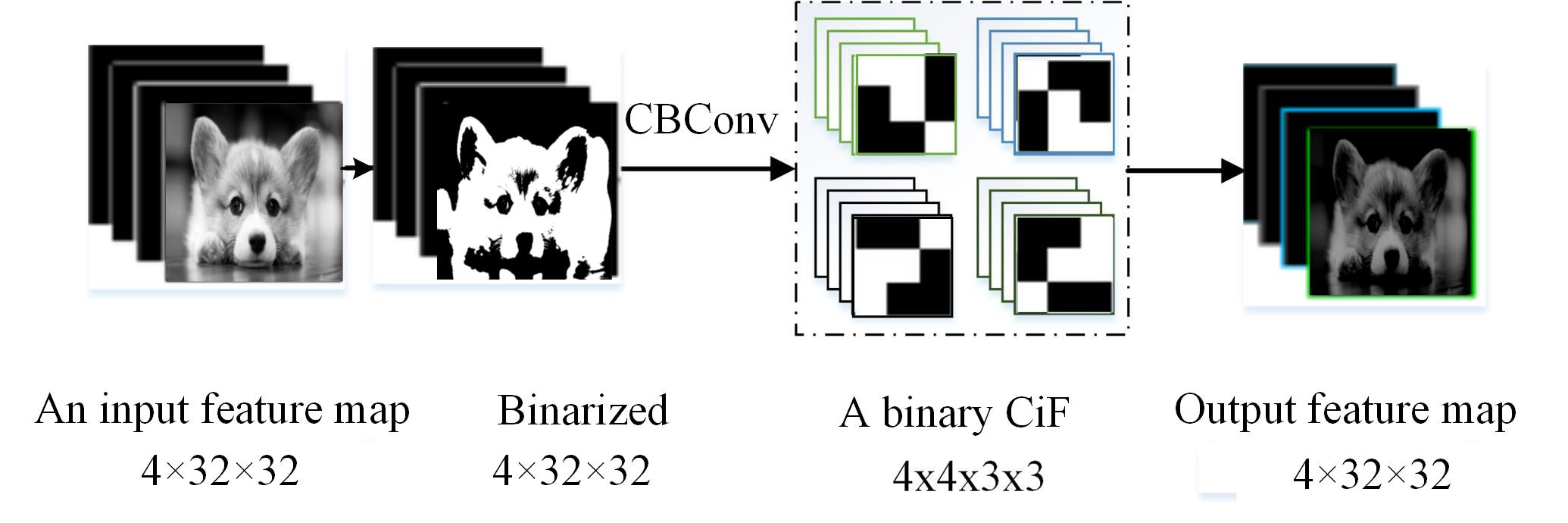}
  }
  \caption{CiF and CBConv examples for $K=4$ orientations ($0^{\circ}$, $90^{\circ}$, $180^{\circ}$, $270^{\circ}$) and $H=3$. (a) The generation of a CiF and its corresponding binary CiF based on a learned filter and $M$. To obtain the $4D$ CiF, the original $2D$ $H\times H$ learned filter is modified to $3D$ by copying it 3 times. (b) CBConv on an input feature map. Note that in this paper, a feature map is defined as $3D$ with $K$ channels, and these channels are usually not the same.}
  \label{4} %
\end{figure}

We now design the specific convolutional filters CiFs used in our CBCNs.
A CiF is a $4D$ tensor of size $K\times K\times H\times H$, generated based on a learned filter and $M$. These CiFs are deployed across all convolutional layers. As shown in Fig. \ref{4_1}, the original $2D$ $H\times H$ learned filter is modified to $3D$ by replicating it three times and concatenating them to obtain the $4D$ CiF. For $K=4$, every channel of the network input is replicated as a group of four elements. By doing so, we can easily implement our CBCNs using PyTorch. One example of the CBConv is illustrated in Fig. \ref{4_2}.

To facilitate the math description below, we represent a $2D$ $H\times H$ learned filter as a $1D$ vector of size $H^2$ so that its corresponding $4D$ CiF can be represented using a $2D$ matrix of size $H^2\times K$ (see Fig. \ref{4_1}). Let $G_i$ be such a matrix representing the $i^{th}$ CiF. Then
\begin{equation}
 \begin{split}
{{G_i}}&=(G_{i0},..., G_{ij},...,G_{i(K-1)})\\
&=(W_{i}\circ m_{0},...,W_{i}\circ m_{j},...,W_{i}\circ m_{K-1}),
 \end{split}
 \label{GiF}
\end{equation}
where $W_i$ is a $1D$ vector containing the $i^{th}$ learned filter's weights (including the  unchanged central one during rotation), and $\circ$ denotes the rotation operation of $W_i$ with $m_j$ (see Fig. \ref{cifmatrix}). $G_{i0}$ corresponds to the $i^{th}$ learned filter and the other columns of $G_i$ are introduced to increase the representational ability.

\subsection{Forward Propagation of CBCNs based on the CBConv Module}
In CBCNs, a binary CiF denoted by $\hat{G}_i$ is calculated as:
\begin{equation}
\hat{G_i} = sign(G_i),
\end{equation}
where $G_i$ is the corresponding full-precision CiF, and the values of $\hat{G}_i$ are 1 or -1. Both $G_i$ and $\hat{G}_i$ are jointly obtained in the end-to-end learning framework. Let the set of all the binarized filters in the $l^{th}$ layer be $\hat{G}^l$. Then the output feature maps $F^{l+1}$ are obtained by:
\begin{equation}
F^{l+1} = CBConv(sign(F^l);\hat{G}^l),
\end{equation}
where $CBConv$ is the convolution operation implemented as a new module including the CiF generation process (the blue part in Fig. \ref{forwardbp}). Fig. \ref{4_2} shows a simple example of a forward convolution where there is one input feature map with one generated output feature map. In the CBConv, the channels of an output feature map are generated as follows:
\begin{equation}
{{F}^{l+1}_{h,j}}=\sum_{i,g} {F_g^l*\hat{G}^l_{ij}},
\end{equation}

\begin{equation}
{{F}^{l+1}_h}=\{{F}^{l+1}_{h,0},{F}^{l+1}_{h,1}, ...,{F}^{l+1}_{h,K-1}\},
\end{equation}
where $*$ denotes the convolution operation. $F^{l+1}_{h,k}$ is the $k^{th}$ channel of the $h^{th}$ feature map, and $F^l_g$ denotes the $g^{th}$ feature map of the input in the $l^{th}$ convolutional layer. In Fig. \ref{4_2}, $h = 1$ and $g = 1$, where after the CBConv with one binary CiF, the number of the channels of the output feature map is the same as that of the input feature map.

\subsection{ Circulant Back Propagation (CBP)}
During the back-propagation, what needs to be learned and updated are the learned filters only. And the inverse transformation of the circulant transfer matrix $M$ is involved in the process of BP to further enhance the representational ability of our CBCNs. To facilitate the math description below, we define the inverse circulant matrix $P$ of $K$ columns as $P=(p_0,...,p_j,...,p_{K-1})$, $j=\{0,1,..,K-1\}$, where $K=8$ and $8$ vector inverse rotations are used to form ${P}$.
Let ${\delta}_{W_i}$ be the gradient of a learned filter $W_i$. Note that we need to sum up the gradients of the $K$ sub-filters
in the corresponding CiF, $G_i$.  Thus:

\begin{equation}
{\delta}_{W_i} =  (\sum_{j=0}^{K-1} \frac{\partial \hat{L}}{\partial G_{ij}(p_j(0))},...,\sum_{j=0}^{K-1} \frac{\partial \hat{L}}{\partial G_{ij}(p_j(7))}),
\label{delta}
\end{equation}
\begin{equation}
W_i \leftarrow W_i - {\eta} {\delta}_{W_{i}},
\end{equation}
where $\hat{L}$ is the network loss function, and $\eta$ is the learning rate.
Note that since the central weights of CiFs are not rotated, their gradients are obtained as in the common BP procedure and are not presented in Eq. \ref{delta}.
As is shown, the circular operation involves in our BP process, which makes CBP be adaptive to orientation variations in objects.

\begin{figure*}
  \centering
  \subfigure[]{
    \label{6_1} %
    \includegraphics[width=2in]{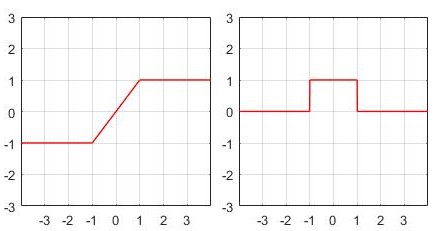}
  }
  \subfigure[]{
    \label{6_2} %
    \includegraphics[width=2in]{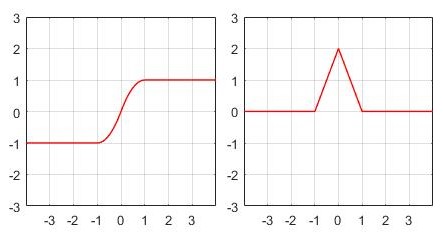}
  }
    \subfigure[]{
    \label{6_3} %
    \includegraphics[width=2in]{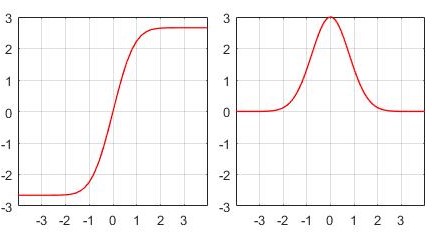}
  }
  \caption{ Three approximations of the sign function for its gradient computation. (a) The clip function and its derivative in \cite{Courbariaux2016Binarized}. (b) The piecewise polynomial function and its derivative in \cite{Liu2018Bi}. (c) Our proposed function and its derivative.}
  \label{6} %
\end{figure*}

For the gradient of the sign function, some special process is necessary due to its discontinuity property. In \cite{Courbariaux2016Binarized} and \cite{Liu2018Bi}, the sign function is approximated by the clip function and the piecewise polynomial function, respectively, as shown in Fig. \ref{6_1} and Fig. \ref{6_2} where their corresponding derivatives are also given. Since the derivative of the sign function (an impulse) can be represented as $\lim\limits_{\sigma\rightarrow 0} \frac{1}{\sigma\sqrt{\pi}}\exp(-\frac{G^2}{\sigma^2})$, in this work, we use this Gaussian function (Fig. \ref{6_3}) as the approximation of the gradient:
\begin{equation}
\frac{\partial \hat{G}_i}{\partial {{G}_i}}=\frac{A}{\sigma\sqrt{\pi}}\exp(-\frac{G_i^2}{\sigma^2}),
\label{gaussian}
\end{equation}
where $A$ and $\sigma$ are the amplitude gain and variance of the Gaussian function, respectively, which are determined empirically. In our experiments, we find that our approximation in Fig. \ref{6_3} is better than those in Fig. \ref{6_1} and Fig. \ref{6_2}. From the equations above, we can see that the BP process can be easily implemented. Thus only updating the learned convolution filters with the help of CiFs, our CBCNs are significantly compact and efficient, reducing the memory storage by 32. Finally, the learning algorithm to train CBCNs is given in Algorithm \ref{algo}.
\begin{algorithm}[htb]
\caption{ CBCN Training.}
\label{algo}
\begin{algorithmic}[1]
\Require
 The training dataset; the full-precision learned filters $W$;
 the circulant transfer matrix $M$;
 the number of orientations $K$;
 hyper-parameters such as initial learning rate, weight decay,
 convolution stride and padding size.
\Ensure
 A CBCN based on the CiFs.
\State Initialize $W$ randomly;
\Repeat
\State // Forward propagation
\ForAll {$l = 1$ to $L$ convolutional layer}
\State Use Eqs. \ref{M} and \ref{GiF} to obtain $G^l$;
\State $F^{l+1} = CBConv$(sign$(F^l)$, sign$(G^l))$;
\EndFor
\State // Back propagation
\ForAll {$l = L$ to $1$}
\State Calculate the gradients ${\delta}_W$;\,//\,Using Eq.\;\ref{delta}
\State  $W \leftarrow W - {\eta} {\delta}_W$; // Update the parameters
\EndFor
\Until{the maximum epoch.}
\end{algorithmic}
\end{algorithm}

\begin{table*}[]
\centering
\caption{ Error rates on the MNIST and CIFAR10 and their variants. `fp' denotes the full precision result. The bold denotes the best result among the binary networks.}
\begin{tabular}{cccccclccclll}
\hline
\multirow{2}{*}{Dataset} & \multirow{2}{*}{Backbone}   & \multirow{2}{*}{kernel stage} & \multicolumn{4}{c}{original\;(\%)}                & \multicolumn{6}{c}{rot\;(\%)}                \\ \cmidrule(lr){4-7}\cmidrule(lr){8-13} %
                          &                           &                               & fp       & XNOR      & \multicolumn{2}{c}{CBCN} & fp    & XNOR   & \multicolumn{4}{c}{CBCN}  \\ \hline\hline
\multirow{2}{*}{MNIST}    & \multirow{2}{*}{LeNet}   & 5-10-20-40                    & 0.91     & 3.76      & \multicolumn{2}{c}{\textbf{1.91}} & 2.77  & 17.26  & \multicolumn{4}{c}{\textbf{5.76}}  \\
 &                           & 10-20-40-80                   & 0.69     & 1.50      & \multicolumn{2}{c}{\textbf{1.24}} & 1.89  & 7.77   & \multicolumn{4}{c}{\textbf{4.95}}  \\\hline
\multirow{3}{*}{CIFAR10}  & \multirow{3}{*}{ResNet18} & 16-16-32-64                   &  {8.94}     & 22.88     & \multicolumn{2}{c}{\textbf{10.9}} & {19.07}      & 40.75  & \multicolumn{4}{c}{\textbf{19.68}} \\
&                           & 32-32-64-128                  &    {6.63}       & 15.55     & \multicolumn{2}{c}{\textbf{8.13}} &  {12.96}    & 33.69  & \multicolumn{4}{c}{\textbf{16.2}} \\
                          &                           & 32-64-128-256                 &  {5.27}         & 13.43     & \multicolumn{2}{c}{\textbf{8.09}}  &{10.47}      & 21.93  & \multicolumn{4}{c}{\textbf{15.11}} \\\hline
\label{rotate}
\end{tabular}
\end{table*}

\section{Experiments}
Our CBCNs are evaluated on  object classification  using MNIST \cite{L1998Gradient}, CIFAR10/100 \cite{Krizhevsky2009Learning} and ILSVRC12 ImageNet datasets \cite{Russakovsky2015ImageNet}.   LeNet \cite{L1998Gradient}, WideResNet (WRN) \cite{Zagoruyko2016Wide} and  ResNet18 \cite{He2016Deep}  are employed as the backbone networks to build our
CBCNs simply by replacing the full-precision convolution  with CBConv. Also,  binarizing the neuron activations is carried out in all of our experiments.
\subsection{Datasets and Implementation Details}
\textbf{Datasets:}  The MINIST \cite{L1998Gradient} dataset is composed of a training set of 60,000 and a testing set of 10,000 $32\times 32$ grayscale images of hand-written digits from 0 to 9.
Each sample is randomly rotated in  $[-45^{\circ}, 45^{\circ}]$  yielding MNIST-rot.

CIFAR10 \cite{Krizhevsky2009Learning} is a natural image classification dataset containing a training set of $50,000$ and a testing set of $10,000$ $32\times 32$ color images across the following 10 classes: airplanes, automobiles, birds, cats, deers, dogs,
frogs, horses, ships, and trucks, while CIFAR100 consists of 100 classes. And we randomly rotate each sample in the CIFAR10 dataset between $[0, 360^{\circ}]$ to yield CIFAR10-rot.

\begin{figure*}[!t]
\centering
\includegraphics[scale=0.4]{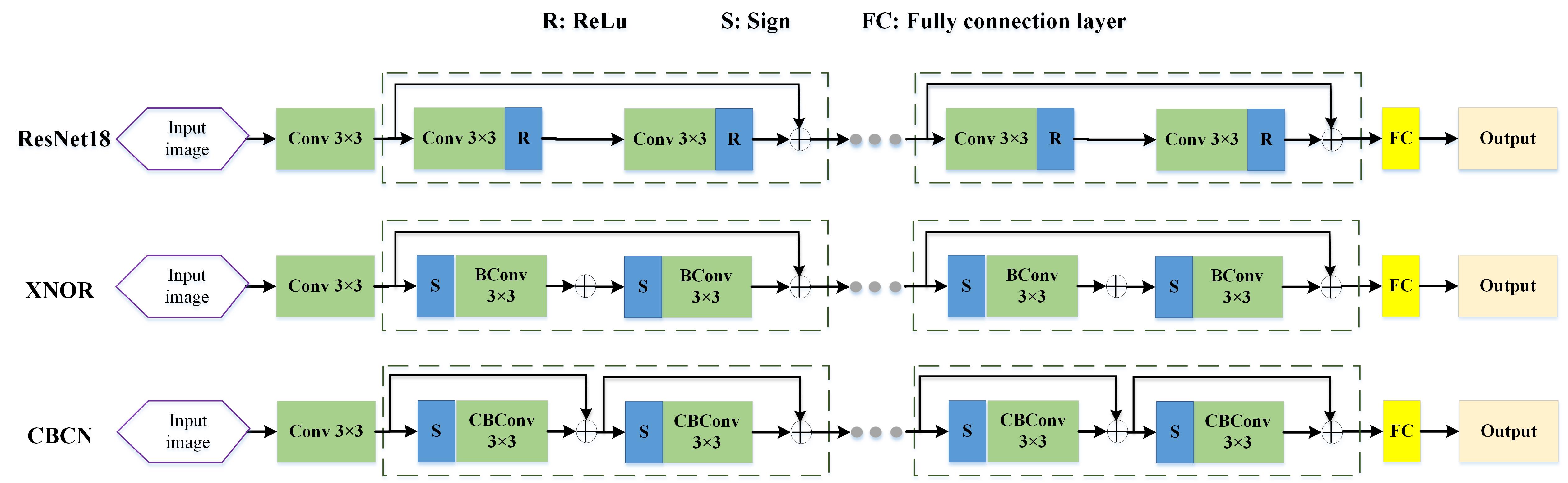}
\caption{Network architectures of ResNet18, XNOR on ResNet18 and CBCN on ResNet18. Note that CBCN doubles the shortcuts.}
\label{architecture}
\end{figure*}
ImageNet object classification dataset \cite{Russakovsky2015ImageNet} is more challenging due to its large scale and greater diversity.
There are 1000 classes and 1.2 million training images and 50k validation images in it. We compare our method with the state-of-the-art on the ImageNet dataset and we adopt ResNet18 to validate the superiority and effectiveness of CBCNs.

In the implementation, LeNet, WRN, and ResNet18 backbone networks are used to build  CBCNs. We simply replace the full-precision convolution with CBConv,  and keep other components unchanged. The parameters $\sigma$ and $A$ for the Gaussian function in the Eq. \ref{gaussian} are  set to 1 and $3\sqrt{2\pi}$, respectively. More details are elaborated below.

\textbf{LeNet Backbone:}
LetNet contains four simple convolutional layers. We adopt Max-pooling and ReLU after each convolution layer, and a dropout layer after the fully connected layer to avoid over-fitting.
The initial learning rate is 0.01 with no degradation  before reaching the maximum epoch of 50 for MNIST and MNIST-rot.

\textbf{WRN Backbone:} WRN is a network structure similar to ResNet with a depth factor $k$ to control the feature map depth dimension expansion through 3 stages, within which the dimensions remain unchanged.
For simplicity we fix the depth factor to 1.
Each WRN has a parameter $i$ which indicates the channel dimension of the first stage and we set it to 16 leading to a network structures $16$-$16$-$32$-$64$. The training details are the same as in \cite{Zagoruyko2016Wide}.
The initial learning rate is 0.01 with a degradation of 10\% for every 60 epochs before it reaches the maximum epoch of 200 for CIFAR10/100 and CIFAR10-rot.
For example, WRN22 is a network with 22 convolutional layers and similarly for WRN40.

\begin{table}[]
\centering
\caption{Performance (accuracy, \%) contributions of the components in CBCNs on CIFAR10, where ConvComp, S, C, and G denote the convolution comparison between BConv in XNOR and CBConv, doubled shortcuts, using the center loss, and using the Gaussian gradient function, respectively. The bold number represents the best result.}
\begin{tabular}{cccccc}
\hline
&\begin{tabular}[c]{@{}l@{}}Conv\\ -Comp\end{tabular} & S & S+C & S+G & \begin{tabular}[c]{@{}l@{}}S+C\\ +G\end{tabular}
\\\hline\hline
 XNOR                  & 76.3      &   80.53      & 80.97& 81.65&\textbf{82.32}\\
CBCN ($K$=2)                  & 81.84        &   85.79      & 86.23& 86.67&\textbf{87.56}\\
CBCN ($K$=4) &84.79 &89.10 &89.6 &90.22 &\textbf{90.83} \\
CBCN ($K$=8)                  & 86.79       &   90.80      &91.27& 91.53&\textbf{92.02}\\ \hline
\label{ablation}
\end{tabular}
\end{table}
\textbf{ResNet18 Backbone:}
Fig. \ref{architecture} respectively illustrates the architectures of ResNet18,  XNOR and CBCNs.
SGD is used as the optimization algorithm with a momentum of $0.9$ and a weight decay 1e-4.
The initial learning rate is 0.01 with a degradation of 10\% for every 20 epochs before reaching the maximum epoch of 70 on ImageNet, while on CIFAR10/100, the initial rate is 0.01 with a degradation of 10\% for every 60 epochs before reaching the maximum epoch of 200.

\subsection{Rotation Invariance}
With LeNet and ResNet18 backbones, we build XNOR and CBCNs and compare them on MNIST, MNIST-rot, CIFAR10, and CIFAR10-rot.  $K$  is set to $4$ in CBCNs.

Table \ref{rotate} gives the results in terms of error rates, and `fp' represents the full-precision results. The state-of-the-art XNOR has a dramatical performance drop on the more challenging rotated datasets. On MNIST-rot, with the kernel stage $5$-$10$-$20$-$40$, CBCN shows impressive performance improvement 11.5\% over XNOR, while 1.85\% improvement is achieved on MNIST. On CIFAR10-rot, with the kernel stage $16$-$16$-$32$-$64$, CBCN has about 20\% improvement over XNOR. From Table \ref{rotate}, we can also see that on CIFAR10-rot, the performance gap between CBCN and XNOR decreases from about \,20\% to 17\% to 6\% with the increase of the kernel stage (parameters), meaning that the improvement of CBCN over XNOR is more significant when they have fewer parameters. The results in Table \ref{rotate} confirm that with the improved representation ability from the proposed CiFs, CBCNs are more robust than conventional binarization methods for rotation variations of input images.

\begin{figure}
  \centering
  \subfigure[Train accuracy on CIFAR10.]{
    \label{loss_cifar1} %
    \includegraphics[width=2.3in]{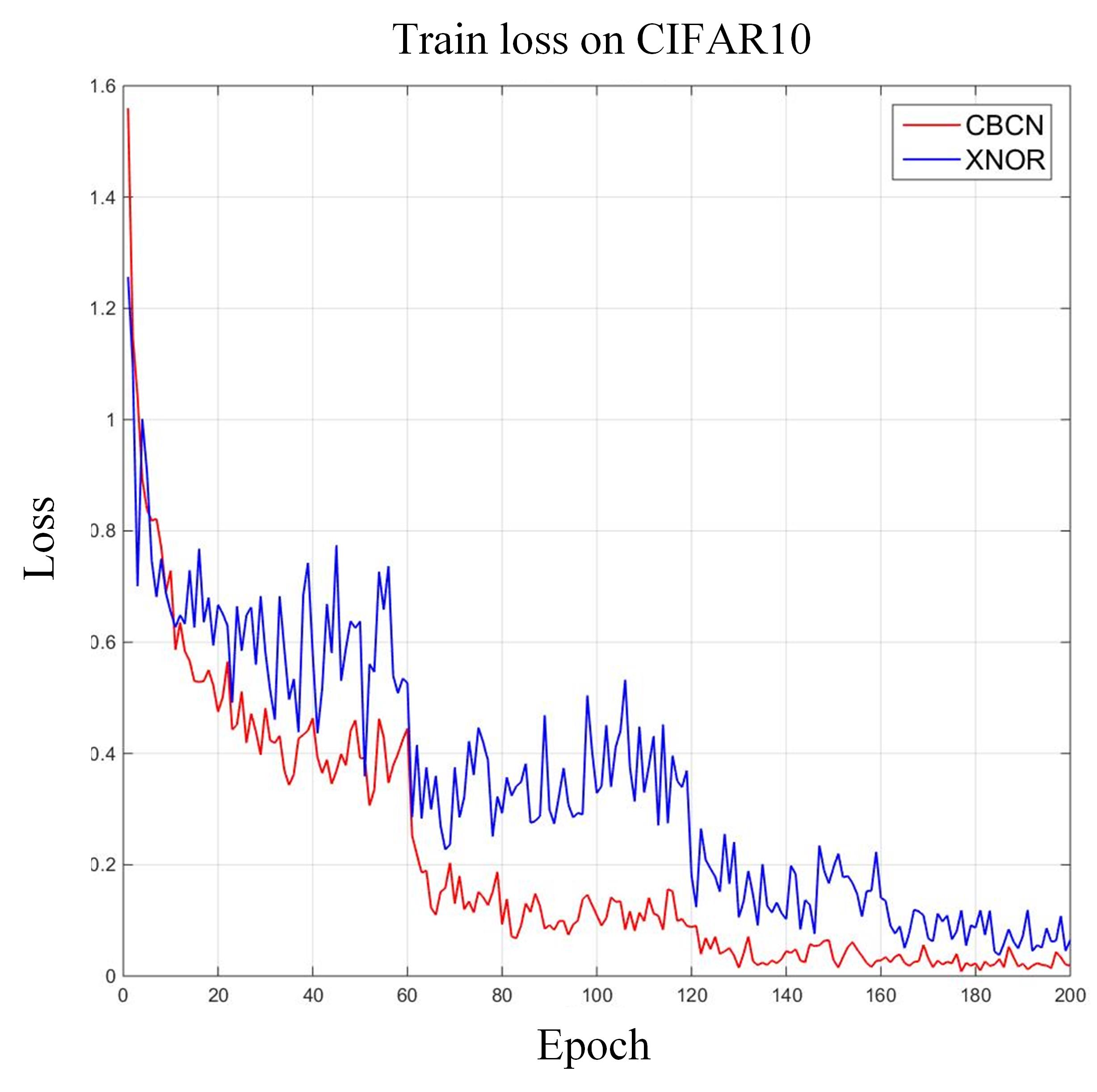}
  }
  \subfigure[Test accuracy on CIFAR10.]{
    \label{loss_cifar2} %
    \includegraphics[width=2.3in]{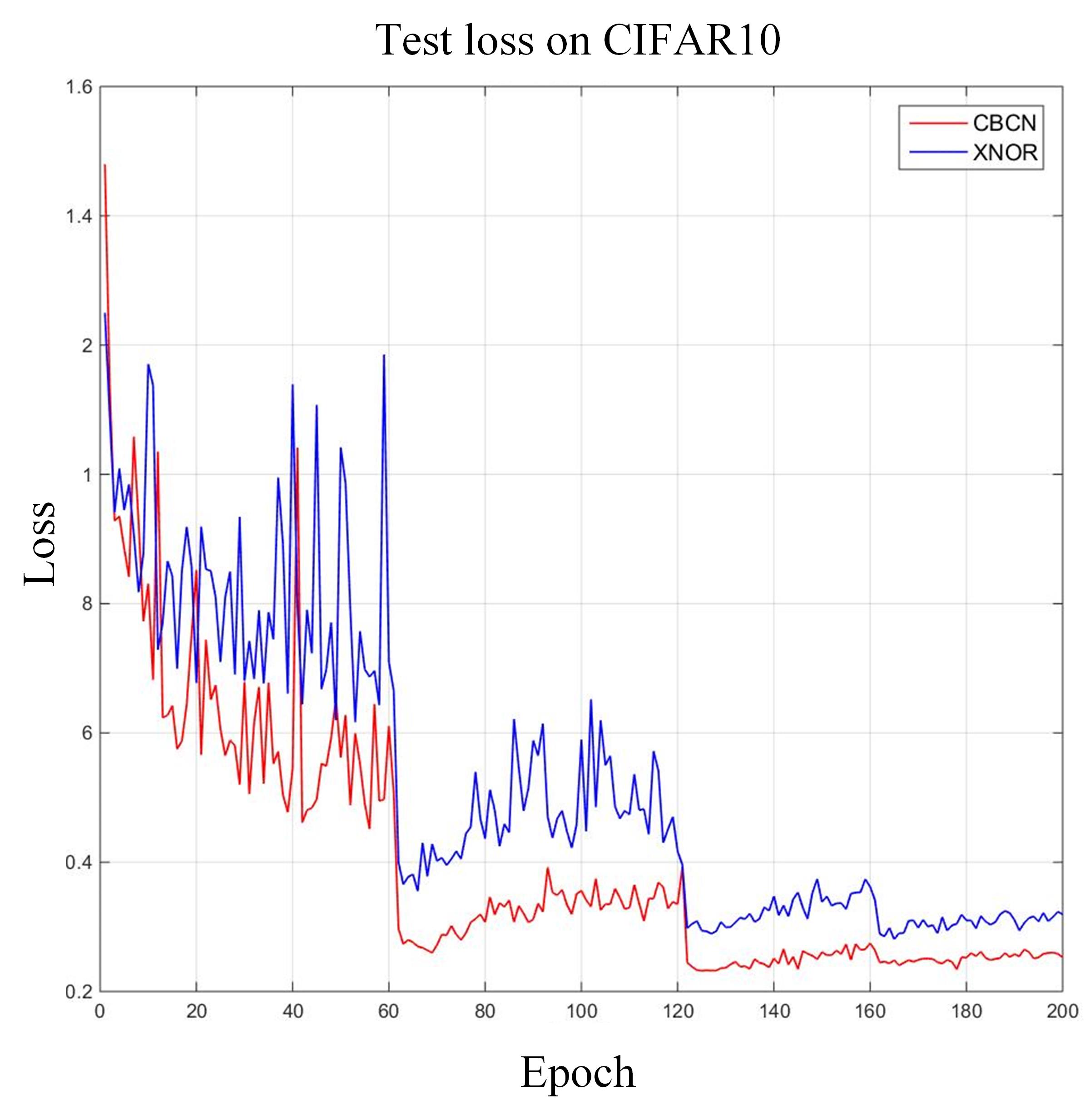}
  }
\caption{Training and Testing error curves of CBCN and XNOR based on WRN40 for the CIFAR10 experiments.}
\label{losscifar}
\end{figure}

\subsection{Ablation Study}
In this section, we study the performance contributions of the components in CBCNs, which include CBConv, center loss, additional shortcuts (Fig. \ref{architecture}), and the Gaussian gradient function (Eq. \ref{gaussian}). CIFAR10 and ResNet18 with kernel stage $16$-$16$-$32$-$64$ are used in this experiment. The details are given below.
\begin{table}[]
\centering
\caption{ Classification accuracy (\%) based on ResNet18 and WRN40, respectively, on CIFAR10/100. The bold  represents the best result among the binary networks. $K=4$ in CBCN.}
\begin{tabular}{cccc}
\hline
\multirow{3}{*}{Model}&\multirow{3}{*}{Kernel Stage}& \multicolumn{2}{c}{Dataset} \\ \cline{3-4}
                      & & CIFAR     & CIFAR   \\
                      & & -10&-100 \\ \hline\hline
BNN              &- & 89.85        & -        \\
BWN                  &- & 90.12         & -        \\
XNOR (ResNet18)                  &64-64-128-256 & 87.1         & 66.08        \\
XNOR (WRN40)                  &64-64-128-256 & 91.58        & 73.18        \\\hline
ResNet18             &16-16-32-64 & 94.84        & 75.37        \\
CBCN                  &16-16-32-64 & 90.22         & 69.97        \\
CBCN                  &32-64-128-256 & \textbf{91.60}         & \textbf{70.07}        \\\hline
WRN40                &16-16-32-64  & 95.8         & 79.41        \\
WRN22                &16-16-32-64  & 90.32         & 67.19        \\
CBCN                 &16-16-32-64& \textbf{93.42}        & \textbf{74.80}        \\ \hline
\end{tabular}
\label{cifartable}
\end{table}

\begin{table*}[]
\centering
\caption{ Classification accuracy (\%) on ImageNet. The bold  represents the best result among the binary networks. $K=4$ in CBCN.}
\begin{tabular}{cccccccc}
\hline
                          &       & Full-Precision & XNOR & ABC-Net & BinaryNet & Bi-Real & CBCN      \\ \hline\hline
\multirow{2}{*}{ResNet18} & Top-1 & 69.3         & 51.2   & 42.7  & 42.2    & 56.4  & \textbf{61.4} \\
                          & Top-5 & 89.2         & 73.2   & 67.6  & 67.1    & 79.5  & \textbf{82.80} \\ \hline
\end{tabular}
\label{imagenettable}
\end{table*}
1) We only replace the convolution BConv in XNOR with our CBConv convolution and compare the results. As shown in the ConvComp column in Table \ref{ablation}, CBCN ($K$=4) achieves about 8\% accuracy improvement over XNOR (84.79\% vs. 76.3\%) using the same network structure and shortcuts as in ResNet18. This significant improvement verifies the effectiveness of our CBConv.

2) In CBCNs, if we double the shortcuts (Fig. \ref{architecture}), we can also find a decent improvement from 84.79\% to 89.10\% (see the column under S in Table \ref{ablation}), which shows that the increase of shortcuts can also enhance binarized deep networks.

3) Fine-tuning CBCN with the center loss can also improve the performance of CBCN by 0.5\% as shown in the column under S+C in Table \ref{ablation}).

4) Replacing the piecewise polynomial function in \cite{Liu2018Bi} with the Gaussian function for back propagation, CBCN obtains 1.12\% improvement (90.22\% vs. 89.10\%), which shows that the gradient function we use is a better choice.

5) From the column under S in Table \ref{ablation}, we can see that CBCN performs better using more orientations in CiFs. More orientations can better deal with the problem of degraded representation caused by network binarization.

\subsection{Accuracy Comparison with  State-of-the-Art}

\textbf{CIFAR10/100:} The same parameter settings are used in CBCNs  on both CIFAR10 and CIFAR100.
We first compare our CBCNs with original ResNet18 with stage kernels as $16$-$16$-$32$-$64$ and $32$-$64$-$128$-$256$,
followed by a comparison with the original WRNs with the initial channel dimension $16$ in Table \ref{cifartable}.
Then, we compare our results with other state-of-the-arts such as BNN \cite{Courbariaux2015BinaryConnect}, BWN \cite{Rastegari2016XNOR}, and XNOR \cite{Rastegari2016XNOR}.
It is observed that at least 1.84\% (= 93.42\%-91.58\%) accuracy improvement is gained with our CBCN, and in other cases, larger margins are achieved.
Also, we plot the training and testing loss curves of XNOR and CBCN, respectively, in Fig. \ref{losscifar}, which clearly show that CBCN (CBP) converges faster than XNOR (BP).

\begin{figure}
  \centering
  \subfigure[Top 1 accuracy on ImageNet.]{
    \label{loss_imagenet1} %
    \includegraphics[width=2.2in]{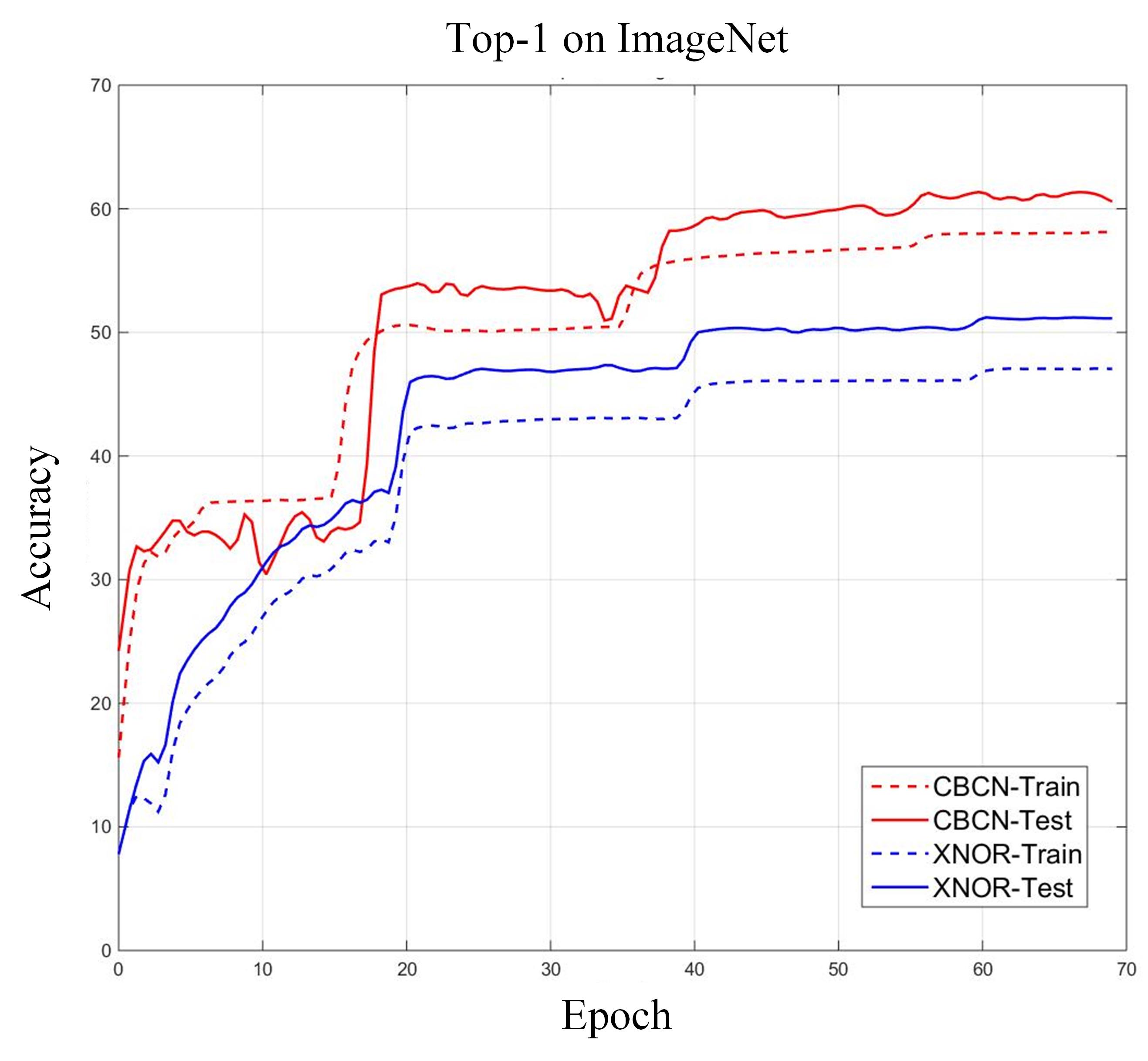}
  }
  \subfigure[Top 5 accuracy on ImageNet.]{
    \label{loss_imagenet2} %
    \includegraphics[width=2.2in]{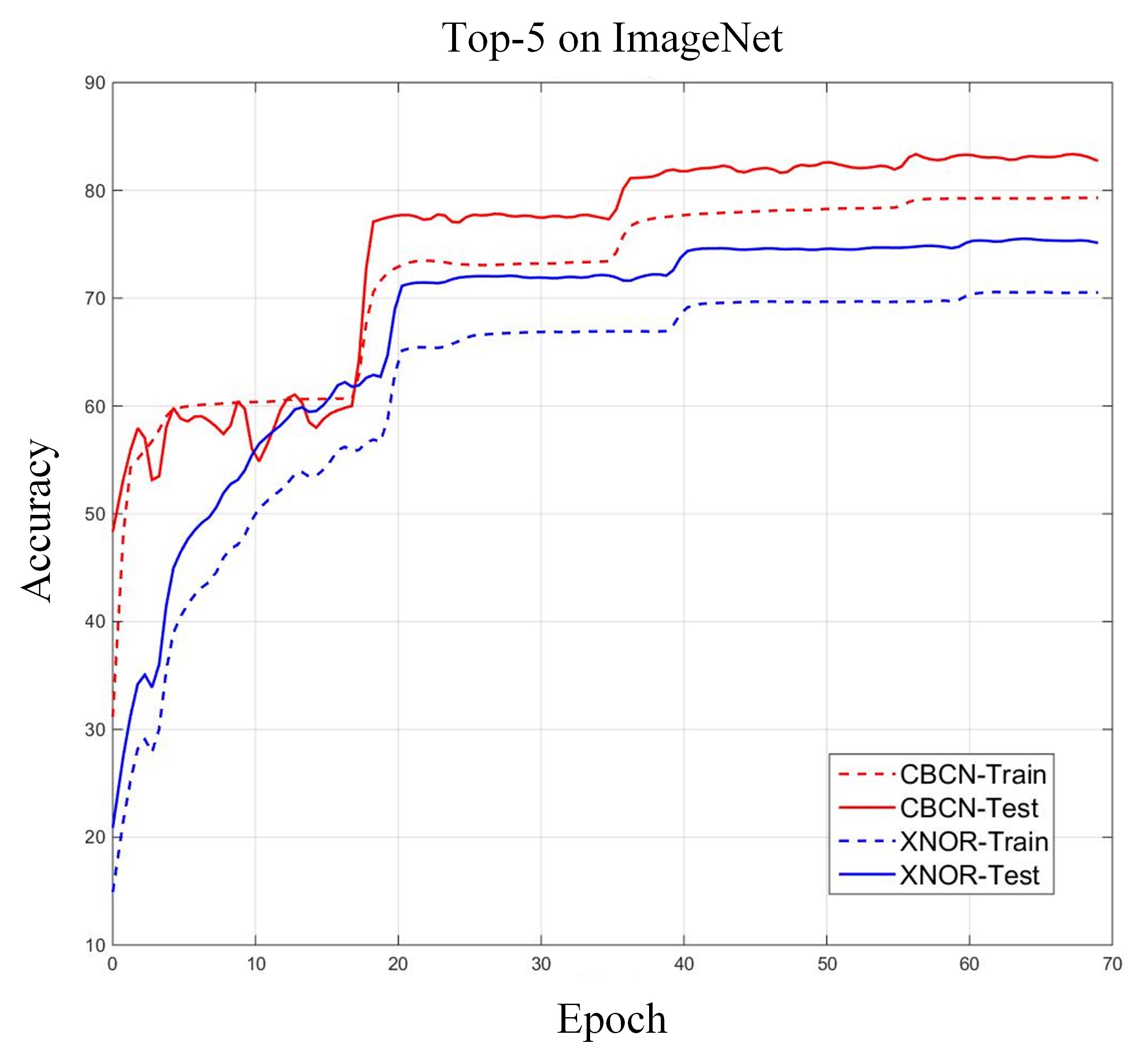}
  }
\caption{Training and Testing error curves of CBCN and XNOR based on the ResNet18 backbone on ImageNet.}
\label{lossimagenet}
\end{figure}

\textbf{ImageNet:} Four  state-of-the-art methods  on ImageNet are chosen for comparison:
Bi-Real Net \cite{Liu2018Bi}, BinaryNet \cite{Courbariaux2016Binarized}, XNOR \cite{Rastegari2016XNOR} and ABC-Net \cite{Lin2017Towards}.
These four networks are representative methods of binarizing both network weights and activations and achieve state-of-the-art results. All the methods in Table \ref{imagenettable} perform the binarization of ResNet18.
For a fair comparison, our CBCN contains the same amount of learned filters as ResNet18.
The comparative results in Table \ref{imagenettable} are quoted directly from the  references, except that the result of BinaryNet is from \cite{Lin2017Towards}.
The comparison clearly indicates that the proposed CBCN outperforms the four binary networks by a considerable margin in terms
of both the top-1 and top-5 accuracies.
Specifically, for top-1 accuracy CBCN outperforms BinaryNet and ABC-Net with a gap over 18\%, achieves about 10\%  improvement over XNOR, and about 5\% over the latest Bi-Real Net.
In Fig. \ref{lossimagenet}, we plot the training and testing loss curves of XNOR and CBCN, respectively.
It clearly shows that using our CBP algorithm, CBCN converges faster than XNOR.

\section{Conclusion}
In this paper, we have proposed new circulant binary convolutional networks (CBCNs) that are implemented by a set of binary circulant filters (CiFs). The proposed CiFs and circulant binary
convolution (CBConv) are used to enhance the representation ability of binary networks. CBCNs can be trained end-to-end with the developed circulant BP (CBP) algorithm. Our extensive experiments demonstrate that CBCNs have superiority over state-of-the-art binary networks, and obtain results that are more close to the full-precision backbone networks ResNets and WRNs, with a storage reduction of about 32 times. As a generic convolutional layer, CBConv can also be used on various tasks, which is our future work.

\section{Acknowledgment}
The work was supported by the National Key Research and Development Program of China
(Grant No. 2016YFB0502602) and the National Key R\&D Plan (2017YFC0821102). Baochang Zhang is the corresponding author.

{\small
\bibliographystyle{ieee}
\bibliography{egbib}
}

\end{document}